\documentclass[10pt,twocolumn,letterpaper]{article}

\usepackage{iccv}
\usepackage{times}
\usepackage{epsfig}
\usepackage{graphicx}
\usepackage{amsmath}
\usepackage{amssymb}
\usepackage[table,dvipsnames]{xcolor}
\usepackage{booktabs}
\usepackage[font=small]{caption}
\usepackage[numbers,sort,compress]{natbib}
\usepackage{balance}
\usepackage[accsupp]{axessibility} %
\usepackage[official]{eurosym}

\definecolor{aliceblue}{rgb}{0.94, 0.97, 1.0}

\definecolor{editcolor}{RGB}{138,43,226}
\definecolor{editcolor2}{RGB}{31,166,181}

\newcommand{\newedits}[1]{#1}
\newcommand{\neweredits}[1]{#1}

\def\sepappendix{0}

\newcommand{\normsmall}[1]{\lVert#1\rVert}

\usepackage[pagebackref=true,breaklinks=true,colorlinks,bookmarks=false]{hyperref}

\iccvfinalcopy %

\begin{document}

\title{Action-Conditioned 3D Human Motion Synthesis with Transformer VAE}

\author{Mathis Petrovich$^{1}$ \quad Michael J. Black$^{2}$ \quad G\"ul Varol$^1$\\
$^{1}$ LIGM, \'Ecole des Ponts, Univ Gustave Eiffel, CNRS, France \\
$^{2}$ Max Planck Institute for Intelligent Systems, T\"{u}bingen, Germany \\
{\tt\small \{mathis.petrovich,gul.varol\}@enpc.fr, black@tue.mpg.de}\\
{\tt\small \url{https://imagine.enpc.fr/~petrovim/actor} }
}

\maketitle

\begin{abstract}

We tackle the problem of action-conditioned
generation of realistic and diverse human motion sequences.
In contrast to methods that complete, or extend, motion sequences,
this task does not require an initial pose or sequence.
Here we learn an action-aware latent representation
for human motions by training a generative variational autoencoder (VAE).
By sampling from this latent space and querying a certain
duration through a series of positional encodings, we synthesize
variable-length motion sequences conditioned
on a categorical action.
Specifically, we design a Transformer-based architecture, \texttt{ACTOR},
for encoding
and decoding a sequence of parametric SMPL human body models
estimated from action recognition datasets.
We evaluate
our approach on the NTU RGB+D, HumanAct12 and UESTC datasets 
and show improvements over the state of the art.
Furthermore, we present two use cases: improving
action recognition through adding our synthesized data
to training, and motion denoising. 
\newedits{Code and models are available on our project page~\cite{projectpage}}.

\end{abstract}
\section{Introduction}
\label{sec:intro}

\begin{figure}
    \centering
    \includegraphics[width=.95\linewidth]{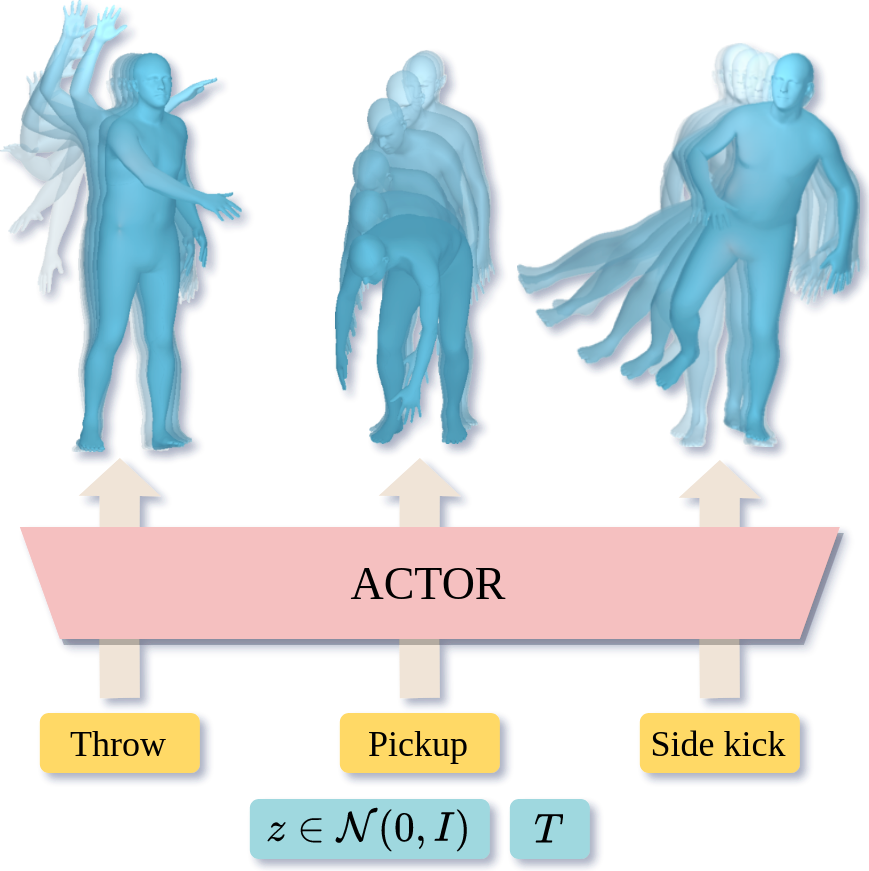}
    \vspace{-0.2cm}
    \renewcommand{\figurename}{Fig.}
    \caption{\textbf{Goal:}
        \texttt{A}ction-\texttt{C}onditioned \texttt{T}ransf\texttt{OR}mer VAE \texttt{(ACTOR)}
        learns to synthesize
        human motion sequences
        conditioned on a categorical action and a duration, $T$.
        Sequences are generated by sampling from a single
        motion representation latent vector, $z$, as opposed to the frame-level
        embedding space in prior work.
    }
    \vspace{-0.6cm}
    \label{fig:teaser}
\end{figure}

Despite decades of research on
modeling human motions \cite{Badler1975thesis,Badler1993}, 
synthesizing realistic and controllable sequences
remains extremely challenging.
In this work, our goal is to take a semantic action label like “Throw”
and generate an infinite number of realistic 3D human
motion sequences, of varying length, that look like realistic throwing (Figure~\ref{fig:teaser}).
A significant amount of prior work has
focused on taking one pose, or a sequence of poses,
and then predicting future motions~\cite{Habibie2017ARV,BarsoumCVPRW2018,Aksan_2019_ICCV,Zhang2020WeAM,yuan2020dlow}.
This is an overly constrained scenario because
it assumes that one already has a motion sequence and just needs more of it.
On the other hand, many applications
such as virtual reality and character control \cite{Holden2017PhasefunctionedNN,Starke2019NeuralSM} 
require generating motions of a given type (semantic action label)
with a specified duration.

We address this problem by training an action-cond\-itioned generative model
with 3D human motion data that has corresponding action labels.
In particular, we construct a Transformer-based encoder-decoder
architecture and train it with the VAE objective.
\neweredits{We parameterize the human body using SMPL~\cite{smpl2015} as it can output joint locations or the body surface. This paves the way for better modeling of interaction with the environment, as the surface is necessary to model contact. Moreover, such a representation allows the use of several reconstruction losses: constraining part rotations in the kinematic tree, joint locations, or surface points. The literature \cite{Lee2018InteractiveCA} and our results suggest that a combination of losses gives the most realistic generated motions.}

The key challenge of motion synthesis
is to generate sequences that are
perceptually realistic while being diverse. 
Many approaches for motion generation have taken an autoregressive approach
such as LSTMs~\cite{Fragkiadaki2015RecurrentNM} and GRUs~\cite{Martinez_2017_CVPR}.
However, these methods typically regress to the mean pose 
after some time~\cite{Martinez_2017_CVPR} and are subject to drift.
The key novelty in our Transformer model is to
provide positional encodings to the decoder
and to output the full sequence at once.
Positional encoding has been popularized by recent work on neural radiance fields \cite{mildenhall2020nerf}; we have not seen it used for motion generation as we do. 
This allows the generation of variable length sequences
without the problem of the motions regressing to the mean pose.
Moreover, our approach is, to our knowledge, the first to create an
action-conditioned \textit{sequence}-level embedding. The closest work is 
Action2Motion~\cite{chuan2020action2motion}, which, in contrast, presents
an autoregressive approach where the latent representation is at the \textit{frame}-level. 
\neweredits{Getting a \textit{sequence}-level embedding requires pooling the time dimension: we introduce a new way of combining Transformers and VAEs for this purpose, which also significantly improves performance over baselines.}

A challenge specific to our action-condition generation problem
is that there exists limited motion capture (MoCap) data
paired with distinct action labels, typically on the order
of 10 categories~\cite{h36m_pami,cmu_mocap}.
We instead rely on monocular motion estimation methods
\cite{VIBECVPR2020} to obtain 3D sequences for actions
and present promising results on 40 fine-grained categories of
the UESTC action recognition dataset~\cite{uestc2018}.
In contrast to~\cite{chuan2020action2motion}, we do not require
multi-view cameras to process monocular trajectory estimates,
which makes our model potentially applicable to larger scales.
Despite being noisy, monocular estimates prove sufficient for training
and, as a side benefit of our model, we are able to
denoise the estimated sequences by encoding-decoding
through our learned motion representation.

An action-conditioned generative model 
can augment 
existing MoCap datasets,
which are expensive and limited in size
\cite{AMASS:ICCV:2019,cmu_mocap}.
Recent work, which renders synthetic
human action videos for training action
recognition models  \cite{varol_surreact}, shows the importance of motion
diversity and large amounts of data per action.
Such approaches
can benefit from an infinite source of
action-conditioned motion synthesis.
We explore this through our experiments on action recognition.
We observe that, despite a domain gap,
the generated motions can 
serve as additional training data, \neweredits{specially} %
in low-data regimes.
Finally, a compact action-aware \neweredits{latent} space for human motions can be used
as a prior in other \neweredits{tasks} %
such as human motion estimation
from videos.

Our contributions are \newedits{fourfold}:
(i) We introduce \mbox{ACTOR},
a novel Transformer-based conditional VAE, and train it to generate action-conditioned human motions by sampling from a sequence-level latent vector.
(ii) We demonstrate that it is possible to learn to generate realistic 3D human motions using noisy 3D body poses estimated from monocular video;
(iii) We present a comprehensive ablation study of the architecture
and loss components, obtaining state-of-the-art performance
on multiple datasets;
(iv) We illustrate two use cases for our model
on action recognition and MoCap denoising.
\newedits{The code is available on our project page~\cite{projectpage}}.
\section{Related Work}
\label{sec:relatedwork}
We briefly review relevant literature on  motion prediction, motion synthesis,
monocular motion estimation, as well as Transformers in the context of VAEs.

\noindent\textbf{Future human motion prediction.}
Research on human motion analysis has a long history
dating back to 1980s~\cite{futrelle1978,ORourke1980ModelbasedIA,Badler1993,Gavrila1999TheVA}.
Given past motion or an initial pose, predicting future frames
has been referred as motion prediction.
Statistical models have been employed in earlier studies~\cite{Bowden2000LearningSM,Galata2001LearningVM}.
Recently, several works show promising results
following progress in generative models with neural networks,
such as GANs~\cite{goodfellow2014gan} or VAEs~\cite{kingma2014auto}.
Examples include HP-GAN~\cite{BarsoumCVPRW2018} and recurrent VAE~\cite{Habibie2017ARV}
for future motion prediction.
Most work treats the body as a skeleton, though recent work exploits full 3D body shape models~\cite{Aksan_2019_ICCV,Zhang2020WeAM}.
Similar to \cite{Zhang2020WeAM}, we also go beyond sparse joints and incorporate vertices on the body surface.
DLow~\cite{yuan2020dlow} focuses on diversifying
the sampling of future motions from a pretrained model.
\cite{Corona2020ContextAwareHM} performs conditional future prediction
using contextual cues about the object interaction.
Very recently, \cite{li2021learn} presents a Transformer-based
method for dance generation conditioned on music and past motion.
Duan et al.~\cite{duan2021singleshot} use Transformers for motion completion.
There is a related line of work on motion ``in-betweening" that takes both past and future poses and ``inpaints" plausible motions between them; see \cite{Harvey:ToG:2020} for more.
In contrast to this prior work, our goal is to
synthesize motions without any past observations.

\noindent\textbf{Human motion synthesis.}
While there is a vast literature on future prediction,
synthesis from scratch has received relatively less attention.
Very early work used PCA \cite{Ormoneit:IVC:2005} and GPLVMs \cite{Urtasun:2007} to learn statistical models of cyclic motions like walking and running.
Conditioning synthesis on multiple, varied, actions is much harder.
DVGANs~\cite{Lin2018HumanMM} %
train a generative model conditioned on a short text
representing actions in MoCap datasets such as Human3.6M~\cite{h36m_pami,IonescuSminchisescu11}
and CMU~\cite{cmu_mocap}.
Text2Action~\cite{Ahn2018Text2ActionGA} 
and Language2Pose~\cite{Ahuja2019Language2PoseNL} similarly
explore conditioning the motion generation on textual descriptions.
Music-to-Dance~\cite{dancing2music2019} and \cite{Li2020LearningTG} study music-conditioned generation.
QuaterNet~\cite{Pavllo2018QuaterNetAQ} focuses on generating locomotion actions
such as walking and running  given a ground trajectory and average speed.
\cite{Yan_2019_ICCV} presents a convolution-based generative
model for realistic, but unconstrained motions without specifying
an action. Similarly, \cite{Zhang2020PerpetualMG} synthesizes
arbitrary sequences, focusing on unbounded motions in time.

\begin{figure*}
    \centering
    \includegraphics[width=.96\linewidth]{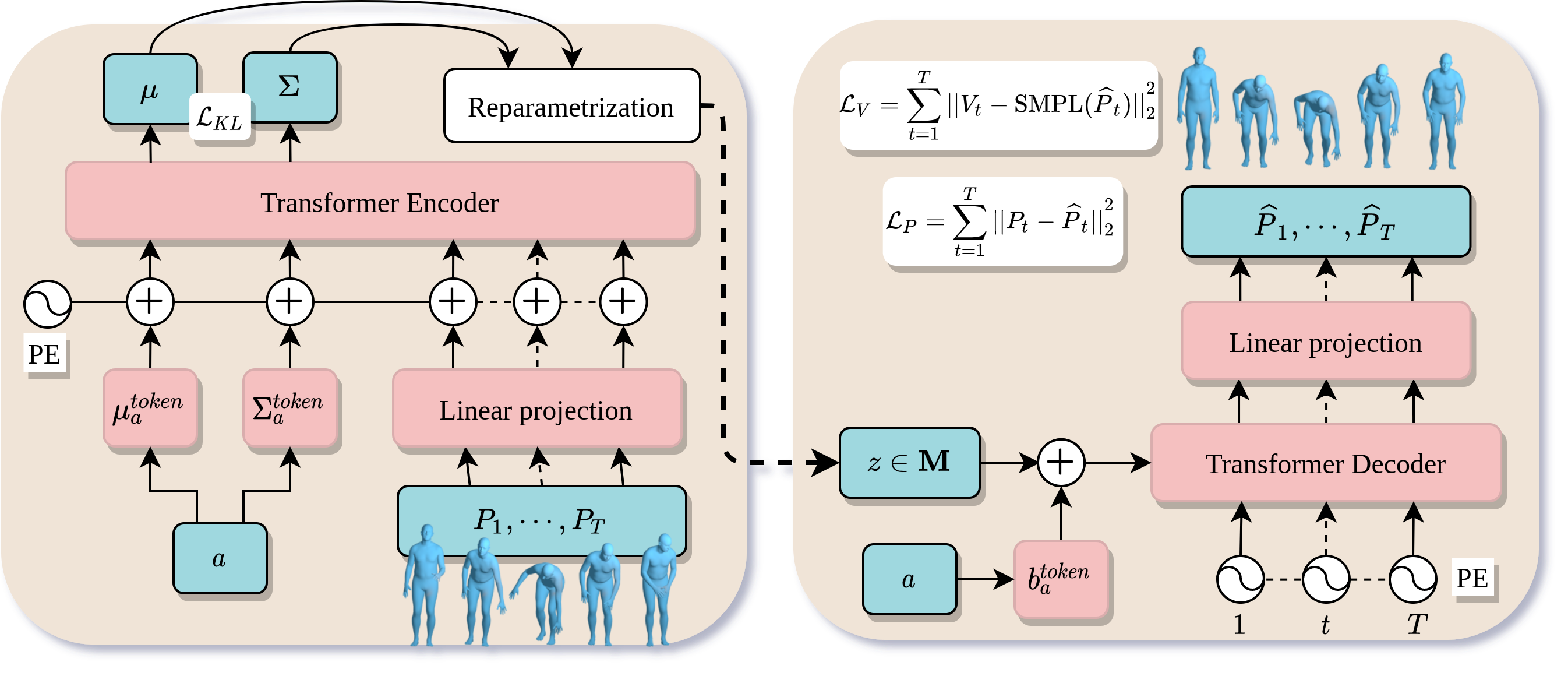}
    \vspace{-0.7cm}
    \caption{\textbf{Method overview:}
        We illustrate the encoder (left) and the decoder (right) of our Transformer-based VAE
        model that generates action-conditioned motions.
        Given a sequence of body poses $P_1, \ldots, P_T$ and an action label $a$, the encoder outputs 
        distribution parameters on which we define a KL loss ($\mathcal{L}_{KL}$).
        We use extra learnable tokens per action
        ($\mu_{a}^{token}$ and $\Sigma_{a}^{token}$) as a way to 
        obtain $\mu$ and $\Sigma$ from the Transformer encoder.
        \neweredits{Using $\mu$ and $\Sigma$, we sample the motion latent representation $z \in \mathbf{M}$.}
        The decoder takes the latent vector $z$, an action label $a$, and a duration $T$ as input.
        The action determines the learnable $b_{a}^{token}$ additive token,
        and the duration determines the number of positional encodings (PE) to input to the decoder.
        The decoder outputs the whole sequence $\widehat{P}_1, \ldots, \widehat{P}_T$ against which
        the reconstruction loss $\mathcal{L}_{P}$ is computed.
        In addition, we compute vertices with a differentiable SMPL layer
        to define a vertex loss ($\mathcal{L}_{V}$).
        For training $z$ is obtained as the output of the encoder; for generation
        it is randomly sampled from a Gaussian distribution.
    }
    \vspace{-0.4cm}
    \label{fig:pipeline}
\end{figure*}

Many methods for unconstrained motion synthesis are
often dominated by actions such as walking and running.
In contrast, our model is able to sample from more general, acyclic, pre-defined action
categories, compatible with action recognition datasets.
In this direction, \cite{Zhao2020BayesianAH} introduces a Bayesian approach,
where Hidden semi-Markov Models are used for jointly training generative
and discriminative models.
Similar to us, \cite{Zhao2020BayesianAH} shows
that their generated motions can serve as additional training data for action recognition. However,
their generated sequences are pseudo-labelled with actions according
to the discriminator classification results.
On the other hand, our conditional model can synthesize motions
in a controlled way, e.g.~balanced training set.
Most similar to our work is
Action2Motion~\cite{chuan2020action2motion}, a \textit{per-frame}
\newedits{VAE}
on actions, using a GRU-based
architecture. Our sequence-level \newedits{VAE} latent space, in conjunction
with the Transformer-based design provides significant advantages,
as shown in our experiments.

\newedits{Other recent works~\cite{weaklynormalizeflowZanfir2020,Henter2020} use
    normalizing flows to address human motion estimation and generation problems.}
\newedits{Several works 
\cite{holden2016motionsynthesis,seong2019denoising,wang2019manifold}
learn a motion manifold,
and use it for motion denoising, which is one of our use cases.}

There is also a significant graphics literature on the topic, which tends to focus on animator control.
See, for example, \cite{Holden:2020} on learning motion matching 
\neweredits{and \cite{Lee2018InteractiveCA} on character animation}. 
Most relevant here are the phase-functioned neural networks \cite{Holden2017PhasefunctionedNN} and neural state machines \cite{Starke2019NeuralSM}.
Both exploit the notion of actions being driven by the phase of a sinusoidal function.
This is related to the idea of positional encoding, but unlike our approach, their methods require manual labor to segment actions and build these phase functions.

\noindent\textbf{Monocular human motion estimation.}
Motion estimation from videos~\cite{humanMotionKZFM19,VIBECVPR2020,Luo20203DHM}
has recently made significant progress but is beyond our scope.
In this work, we adopt VIBE~\cite{VIBECVPR2020} to obtain training motion sequences from action-labelled video datasets.

\noindent\textbf{Transformer VAEs.}
Recent successes of Transformers in language tasks
has increased interest in attention-based
neural network models. Several works use Transformers
in conjunction with generative VAE training.
Particular examples include
story generation~\cite{fang2021transformerbased},
sentiment analysis~\cite{cheng2019variational},
response generation~\cite{lin2020variational}, and
music generation~\cite{jian20_tvae}.
The work of \cite{jian20_tvae} learns latent embeddings
per timeframe, while \cite{cheng2019variational}
averages the hidden states to obtain a single latent code.
On the other hand, \cite{fang2021transformerbased}
performs attention averaging to pool over time.
In contrast to these works, we adopt learnable tokens
as in~\cite{devlin2018bert,dosovitskiy2020image} to summarize the input into a sequence-level
embedding.

\section{Action-Conditioned Motion Generation}
\label{sec:method}

\noindent\textbf{Problem definition.} Actions defined by body-motions can be characterized
by the rotations of body parts, independent of identity-specific body shape.
To be able to generate motions with actors of different morphology, it is desirable to disentangle the pose
and the shape.
Consequently, without loss of generality, we employ the SMPL body model~\cite{smpl2015}, which is a disentangled body representation \newedits{(similar to recent models \cite{MANO:SIGGRAPHASIA:2017,STAR:2020,SMPL-X:2019,ghum2020cvpr}).}
Ignoring shape, our goal, is then to generate a sequence of \textit{pose} parameters.
More formally, given an action label $a$ (from a set of predefined
action categories $a \in A$) and a duration $T$, we 
generate a sequence of body poses $R_1, \ldots, R_T$
and a sequence of translations of the root joint represented as displacements, $D_1, \ldots, D_T$ (with $D_t \in \mathbb{R}^{3}, \forall t \in \{1, \ldots, T\}$).

\noindent\textbf{Motion representation.} SMPL pose parameters per-frame
represent 23 joint rotations in the kinematic tree and one global rotation. We adopt
the continuous 6D rotation representation for training~\cite{Zhou2019OnTC}, making $R_t \in \mathbb{R}^{24\times6}$.
Let $P_t$
be the combination of $R_t$ and $D_t$, representing the pose and location of the body in a single frame, $t$. 
The full motion is the sequence $P_1, \dots, P_T$.
Given a generator output pose $P_t$ and any shape parameter, we can obtain body mesh vertices ($V_t$) and body joint coordinates ($J_t$) differentiably using~\cite{smpl2015}.

\subsection{Conditional Transformer VAE for Motions}
\label{subsec:ctvae}

We employ a conditional variational autoencoder (CVAE) model \cite{NIPS2015_8d55a249}
and input the action category information to both the encoder and the decoder.
More specifically, our model is an action-conditioned Transformer VAE (\mbox{ACTOR}),
whose encoder and decoder consist of Transformer layers
(see Figure \ref{fig:pipeline} for an overview).

\noindent\textbf{Encoder.} The encoder takes an arbitrary-length sequence of poses,
and an action label $a$ as input, and outputs distribution parameters $\mu$ and $\Sigma$ of the motion latent space. 
Using the reparameterization trick~\cite{kingma2014auto}, we sample from this distribution a latent vector
$z \in \mathbf{M}$ with $\mathbf{M} \subset \mathbb{R}^d$.
All the input pose parameters ($R$) and translations ($D$) are first linearly embedded into a $\mathbb{R}^d$ space. As we embed arbitrary-length sequences into one latent space (sequence-level embedding), we need to pool the temporal dimension. 
In other domains, a \verb+[class]+ token has been introduced
for pooling purposes, e.g., in NLP with BERT~\cite{devlin2018bert} and more recently in computer vision with ViT~\cite{dosovitskiy2020image}. %
Inspired by this approach,
\newedits{we similarly prepend the inputs with learnable tokens, and only use the corresponding encoder outputs as a way to pool the time dimension.
To this end,} we include two extra learnable parameters per action, $\mu_a^{token}$ and $\Sigma_a^{token}$, which
\newedits{we called
``distribution parameter tokens''}. 
We append the embedded pose sequences to these tokens.
The resulting Transformer encoder input is the summation with
the positional encodings in the form
of sinusoidal functions.
We obtain the distribution parameters $\mu$ and $\Sigma$ by taking the first two outputs of the encoder corresponding to
the distribution parameter tokens (i.e., discarding the rest).

\noindent\textbf{Decoder.}
Given a single latent vector $z$ and an action label $a$, the decoder generates a realistic human motion for a given duration \newedits{in \neweredits{one shot} (i.e., not autoregressive).} %

We use a Transformer decoder model where we feed time information as a query (in the form of $T$ sinusoidal positional encodings), and the latent vector combined with action information, as key and value.
To incorporate the action information, we simply add a learnable bias $b_a^{token}$ to shift the latent representation to an action-dependent space.
The Transformer decoder outputs a sequence of $T$ vectors in $\mathbb{R}^d$ from which we obtain the final poses $\widehat{P}_1, \ldots, \widehat{P}_T$ following a linear projection.
A differentiable SMPL layer is used to obtain
vertices and joints given the pose parameters as output by the decoder.

\subsection{Training}
\label{subsec:training}
We define several loss terms to train our model
and present an ablation study in Section~\ref{subsec:ablations}.

\noindent\textbf{Reconstruction loss on pose parameters ($\mathcal{L}_{P}$).}
We use an L2 loss between the ground-truth poses $P_1, \ldots, P_T$, and our predictions $\widehat{P}_1, \ldots, \widehat{P}_T$ as $\mathcal{L}_{P} = \sum_{t=1}^T \normsmall{ P_t - \widehat{P}_t }^2_2$. Note that this loss contains both the SMPL rotations and the root translations.
When we experiment by discarding the translations,
we break this term into two: $\mathcal{L}_{R}$ and $\mathcal{L}_{D}$,
for rotations and translations, respectively.

\noindent\textbf{Reconstruction loss on vertex coordinates ($\mathcal{L}_{V}$).}
We feed the SMPL poses $P_t$ and $\widehat{P}_t$ to a differentiable SMPL layer (without learnable parameters) with a mean shape (i.e., $\beta = \vec{0}$) to obtain the root-centered vertices of the mesh $V_t$ and $\widehat{V}_t$.
We define an L2 loss by comparing to the ground-truth vertices ${V}_t$ as
$\mathcal{L}_{V} = \sum_{t=1}^T \normsmall{ V_t - \widehat{V}_t }^2_2$ .
We further experiment with a loss $\mathcal{L}_{J}$
on a more sparse set of points
such as joint locations $\widehat{J}_t$
obtained through the SMPL joint regressor.
However, as will be shown in Section~\ref{subsec:ablations},
we do not include this term in the final model. 

\noindent\textbf{KL loss ($\mathcal{L}_{KL}$).} As in a standard VAE, we 
regularize the latent space by encouraging it to be similar to a Gaussian 
distribution \neweredits{with $\mu$ the null vector and $\Sigma$ the identity matrix}. 
We minimize the Kullback–Leibler (KL) divergence
between the encoder distribution and \newedits{this target distribution.}%

The resulting total loss is defined as the summation of different terms: $\mathcal{L} = \mathcal{L}_{P} + \mathcal{L}_{V} + \lambda_{KL}\mathcal{L}_{KL}$.
We empirically show the importance of weighting with $\lambda_{KL}$ \newedits{(equivalent to the $\beta$ term in $\beta$-VAE \cite{Higgins2017betaVAELB})}
in our experiments
to obtain a good trade-off between diversity and realism (see
\if\sepappendix1{Section~A.1}
\else{Section~\ref{app:subsec:kl}}
\fi
of the appendix).
The remaining loss terms 
are simply equally weighed, further improvements are potentially possible with tuning.
We use the AdamW optimizer with a fixed learning rate of 0.0001.
The minibatch size is set to 20 and we found that the performance is sensitive to this
hyperparameter (see
\if\sepappendix1{Section~A.2}
\else{Section~\ref{app:subsec:batchsize}}
\fi
of the appendix).
We train our model for 2000, 5000 and 1000 epochs on NTU-13, HumanAct12 and UESTC datasets, respectively.
Overall, more epochs produce improved performance, but we stop training
to retain a low computational cost.
Note that
to allow faster iterations, for ablations on loss and architecture,
we train our models for 1000 epochs on NTU-13 and 500 epochs on UESTC.
The remaining implementation details can be found
in 
\if\sepappendix1{Section~C}
\else{Section~\ref{app:sec:implementation}}
\fi
of the appendix.

\section{Experiments}
\label{sec:experiments}

\begin{table*}
\centering
\setlength{\tabcolsep}{6pt}
\resizebox{.99\linewidth}{!}{
\begin{tabular}{lccrcc|cccc}
    \toprule
        & \multicolumn{5}{c}{UESTC} & \multicolumn{4}{|c}{NTU-13} \\
    Loss & FID$_{tr}$$\downarrow$ & FID$_{test}$$\downarrow$ & \multicolumn{1}{c}{Acc.$\uparrow$} & Div.$\rightarrow$ & Multimod.$\rightarrow$ & FID$_{tr}$$\downarrow$ & Acc.$\uparrow$ & Div.$\rightarrow$ & Multimod.$\rightarrow$ \\
        \midrule
          Real & $2.93^{\pm0.26}$ & $2.79^{\pm0.29}$ & $98.8^{\pm0.1}$ & $33.34^{\pm0.32}$ & $14.16^{\pm0.06}$ & $0.02^{\pm0.00}$ & $99.8^{\pm0.0}$ & $7.07^{\pm0.02}$ & $2.27^{\pm0.01}$\\
        \midrule
   $\mathcal{L}_{J}$ & $3$M$^*$ & $3$M$^*$ & $3.3^{\pm0.2}$ & $267.68^{\pm346.06}$ & $153.62^{\pm50.62}$ & $0.49^{\pm0.00}$ \
& $93.6^{\pm0.2}$ & $7.04^{\pm0.04}$ & $2.12^{\pm0.01}$\\
$\mathcal{L}_{R}$ & $292.54^{\pm113.35}$ & $316.29^{\pm26.05}$ & $42.4^{\pm1.7}$ & $23.16^{\pm0.47}$ & $14.37^{\pm0.08}$ & $0.23^{\pm0.00}$ & $95.4^{\pm0.2}$ & $7.08^{\pm0\
.04}$ & $2.18^{\pm0.02}$\\
$\mathcal{L}_{V}$ & $4$M$^*$ & $4$M$^*$ & $2.7^{\pm0.2}$ & $314.66^{\pm476.18}$ & $169.49^{\pm27.90}$ & $0.25^{\pm0.00}$ & $95.8^{\
\pm0.3}$ & $7.08^{\pm0.04}$ & $2.07^{\pm0.01}$\\
\rowcolor{aliceblue}
$\mathcal{L}_{R}$ + $\mathcal{L}_{V}$ & ${\bf 20.49}^{\pm2.31}$ & ${\bf 23.43}^{\pm2.20}$ & ${\bf 91.1}^{\pm0.3}$ & $31.96^{\pm0.36}$ & $14.66^{\pm0.03}$ & ${\bf 0.19}^{\pm0.00}$ & ${\bf 96.2}^{\pm0.2}\
$ & $7.09^{\pm0.04}$ & $2.08^{\pm0.01}$\\
        \bottomrule
\end{tabular}
}
\vspace{-0.3cm}
\caption{\textbf{Reconstruction loss:}
We define the loss on the SMPL pose parameters
which represent the rotations in the kinematic tree ($\mathcal{L}_{R}$), their joint coordinates ($\mathcal{L}_{J}$),
as well as vertex coordinates ($\mathcal{L}_{V}$).
We show that constraining both rotations and vertex coordinates
is critical to obtain smooth motions. In particular, coordinate-based losses alone
do not converge to a meaningful solution on UESTC \neweredits{(*)}.
$\rightarrow$ means motions are better when the metric is closer to real. 
}
\vspace{-0.3cm}
\label{tab:loss}
\end{table*}

We first introduce the datasets and performance measures
used in our experiments (Section~\ref{subsec:datasets}).
Next, we present an ablation study (Section~\ref{subsec:ablations})
and compare to previous work (Section~\ref{subsec:sota}).
Then, we illustrate use cases in
action recognition (Sections~\ref{subsec:usecases}).
Finally, we provide qualitative results
and discuss limitations (Section~\ref{subsec:qualitative}).

\subsection{Datasets and evaluation metrics}
\label{subsec:datasets}
We use three datasets originally proposed for
action recognition, mainly for skeleton-based inputs.
Each dataset is temporally trimmed around one action per sequence.
Next, we briefly describe them.

\noindent\textbf{NTU RGB+D dataset~\cite{NTURGBD,Liu_2019_NTURGBD120}.}
To be able to compare to the work of \cite{chuan2020action2motion},
we use their subset of 13 action categories. \cite{chuan2020action2motion}
provides SMPL parameters obtained through VIBE estimations.
Their 3D root translations, obtained through multi-view constraints,
are not publicly available, therefore we use their approximately
origin-centered version.
We refer to this data as NTU-13
and use it for training.

\noindent\textbf{HumanAct12 dataset~\cite{chuan2020action2motion}.}
Similarly, we use this data for state-of-the-art comparison.
HumanAct12 is adapted from the PHSPD dataset~\cite{zou2020detailed} that releases SMPL pose
parameters and root translations in camera coordinates for 1191 videos.
HumanAct12 temporally trims the videos, annotates them into
12 action categories, and only provides their joint
coordinates in a canonical frame.
We also process the SMPL poses to align them to the frontal view.

\noindent\textbf{UESTC dataset~\cite{uestc2018}.}
This recent dataset consists of 25K sequences
across 40 action categories (mostly exercises, and some represent \neweredits{cyclic} movements).
To obtain SMPL sequences,
we apply VIBE on each video and select the person track
that corresponds best to the Kinect skeleton provided
in case there are multiple people. We use all 8 static viewpoints
(we discard the rotating camera) and canonicalize all bodies
to the frontal view. We use the official cross-subject protocol
to separate train and test splits, instead of the cross-view protocols
since generating different viewpoints is trivial for our model.
This results in 10650 training sequences that we use for
learning the generative model, as well as the recognition model\newedits{: the effective diversity of this set can be seen as 33 sequences per action on average (10K divided by 8 views, 40 actions).}
The remaining 13350 sequences are used for testing.
Since the protocols on NTU-13 and HumanAct12 do not provide test splits,
we rely on UESTC for recognition experiments.

\noindent\textbf{Evaluation metrics.}
We follow the performance measures
employed in \cite{chuan2020action2motion}
for quantitative evaluations.
We measure FID, action recognition accuracy,
overall diversity, and per-action diversity
(referred to as multimodality in \cite{chuan2020action2motion}).
For all these metrics, a pretrained action recognition
model is used, either for extracting motion features
to compute FID, diversity, and multimodality; or
directly the accuracy of recognition. For experiments on NTU-13 and HumanAct12, we
directly use the provided recognition models of \cite{chuan2020action2motion}
that operate on joint coordinates.
For UESTC, we train our own recognition model
based on pose parameters expressed as 6D rotations
(we observed that the joint-based models of \cite{chuan2020action2motion}
are sensitive to global viewpoint changes).
We generate sets of sequences 20 times with different
random seeds and report the average together with the confidence interval at 95\%.
We refer to
\cite{chuan2020action2motion} for further details.
One difference in our evaluation is the use of average
shape parameter ($\beta=\vec{0}$) when obtaining joint coordinates
from the mesh for both real and generated sequences.
Note also that \cite{chuan2020action2motion} only reports
FID score comparing to the training split (FID$_{tr}$),
since NTU-13 and HumanAct12 datasets do not provide
test splits. On UESTC, we additionally provide an FID score
on the test split as FID$_{test}$, which we rely most on
to make conclusions.

\begin{table*}
    \centering
    \setlength{\tabcolsep}{6pt}
    \resizebox{0.99\linewidth}{!}{
    \begin{tabular}{lccccc|cccc}
        \toprule
        & \multicolumn{5}{c}{UESTC} & \multicolumn{4}{|c}{NTU-13} \\
        Architecture & FID$_{tr}$$\downarrow$ & FID$_{test}$$\downarrow$ & Acc.$\uparrow$ & Div.$\rightarrow$ & Multimod.$\rightarrow$ & FID$_{tr}$$\downarrow$ & Acc.$\uparrow$ & Div.$\rightarrow$ & Multimod.$\rightarrow$ \\
        \midrule
        Real & $2.93^{\pm0.26}$ & $2.79^{\pm0.29}$ & $98.8^{\pm0.1}$ & $33.34^{\pm0.32}$ & $14.16^{\pm0.06}$ & $0.02^{\pm0.00}$ & $99.8^{\pm0.0}$ & $7.07^{\pm0.02}$ & $2.27^{\pm0.01}$\\
        \midrule
        Fully connected & $562.09^{\pm48.12}$ & $548.13^{\pm38.34}$ & $10.5^{\pm0.5}$ & $12.96^{\pm0.11}$ & $10.87^{\pm0.05}$ & $0.47^{\pm0.00}$ & $88.7^{\pm0.6}$ & $6.93^{\pm0.03}$ & $3.05^{\pm0.01}$\\
        GRU & $25.96^{\pm3.02}$ & $27.08^{\pm2.98}$ & $87.3^{\pm0.4}$ & $30.66^{\pm0.33}$ & $15.24^{\pm0.08}$ & $0.28^{\pm0.00}$ & $94.8^{\pm0.2}$ & $7.08^{\pm0.04}$ & $2.20^{\pm0.01}$\\
        \midrule
        \rowcolor{aliceblue}
        Transformer & ${\bf 20.49}^{\pm2.31}$ &  ${\bf 23.43}^{\pm2.20}$ & ${\bf 91.1}^{\pm0.3}$ & $31.96^{\pm0.36}$ & $14.66^{\pm0.03}$ & $0.19^{\pm0.00}$ & ${\bf 96.2}^{\pm0.2}$ & $7.09^{\pm0.04}$ & $2.08^{\pm0.01}$\\
        \quad \newedits{a) w/ autoreg. decoder} & $55.75^{\pm2.62}$ & $60.10^{\pm4.87}$ & $88.4^{\pm0.6}$ & $33.46^{\pm0.69}$ & $10.62^{\pm0.10}$ & $2.62^{\pm0.01}$ & $88.0^{\pm0.5}$ & $6.80^{\pm0.03}$ & $1.76^{\pm0.01}$\\
        
        \quad b) w/out $\mu_{a}^{token},\Sigma_{a}^{token}$ & $27.46^{\pm3.43}$ & $31.37^{\pm3.04}$ & $86.2^{\pm0.4}$ & $31.82^{\pm0.38}$ & $15.71^{\pm0.12}$ & $0.26^{\pm0.00}$ & $94.7^{\pm0.2}$ & $7.09^{\pm0.03}$ & $2.15^{\pm0.01}$\\
        \quad c) w/out $b_{a}^{token}$ & $24.38^{\pm2.37}$ & $28.52^{\pm2.55}$ & $89.4^{\pm0.7}$ & $32.11^{\pm0.33}$ & $14.52^{\pm0.09}$ & ${\bf 0.16}^{\pm0.00}$ & ${\bf 96.2}^{\pm0.2}$ & $7.08^{\pm0.04}$ & $2.19^{\pm0.02}$\\
        \bottomrule
    \end{tabular}
    }
    \vspace{-0.3cm}
    \caption{\textbf{Architecture:}
        We compare various architectural designs,
        such as the encoder and the decoder of the VAE,
        and different components of the Transformer
        model,
        on both NTU-13 and UESTC datasets.
    }
    \vspace{-0.3cm}
    \label{tab:arch}
\end{table*}

\subsection{Ablation study}
\label{subsec:ablations}
We first ablate several components of our approach
in a controlled setup, studying the loss
and the architecture.

\noindent\textbf{Loss study.}
Here, we investigate the influence of the reconstruction
loss formulation when using the parametric SMPL body model
in our VAE.
We first experiment with using (i) only the rotation parameters $\mathcal{L}_{R}$,
(ii) only the joint coordinates $\mathcal{L}_{J}$,
(iii) only the vertex coordinates $\mathcal{L}_{V}$,
and (iv) the combination $\mathcal{L}_{R} + \mathcal{L}_{V}$.
Here, we initially discard the root translation to only assess the pose representation.
Note that for representing the rotation parameters, we use
the 6D representation from~\cite{Zhou2019OnTC}
(further studies on losses with different rotation representations
can be found in
\if\sepappendix1{Section~A.4}
\else{Section~\ref{app:subsec:poserep}}
\fi
of the appendix).
In Table~\ref{tab:loss},
we observe that a single loss is not sufficient
to constrain the problem, especially losses on the coordinates
do not converge to a meaningful solution on UESTC.
On NTU-13, qualitatively,
\newedits{we also observe invalid body shapes since joint locations
alone do not fully constrain the rotations along limb axes.}
We provide examples in our qualitative analysis.
We conclude that using a combined loss significantly
improves the results, constraining the pose space more effectively.
We further provide an experiment on the influence
of the weight parameter $\lambda_{KL}$ controlling
the KL divergence loss term $\mathcal{L}_{KL}$ 
in 
\if\sepappendix1{Section~A.1}
\else{Section~\ref{app:subsec:kl}}
\fi
of the appendix
and note its importance
to obtain high diversity performance.

\begin{figure}
    \centering
    \includegraphics[width=0.99\linewidth]{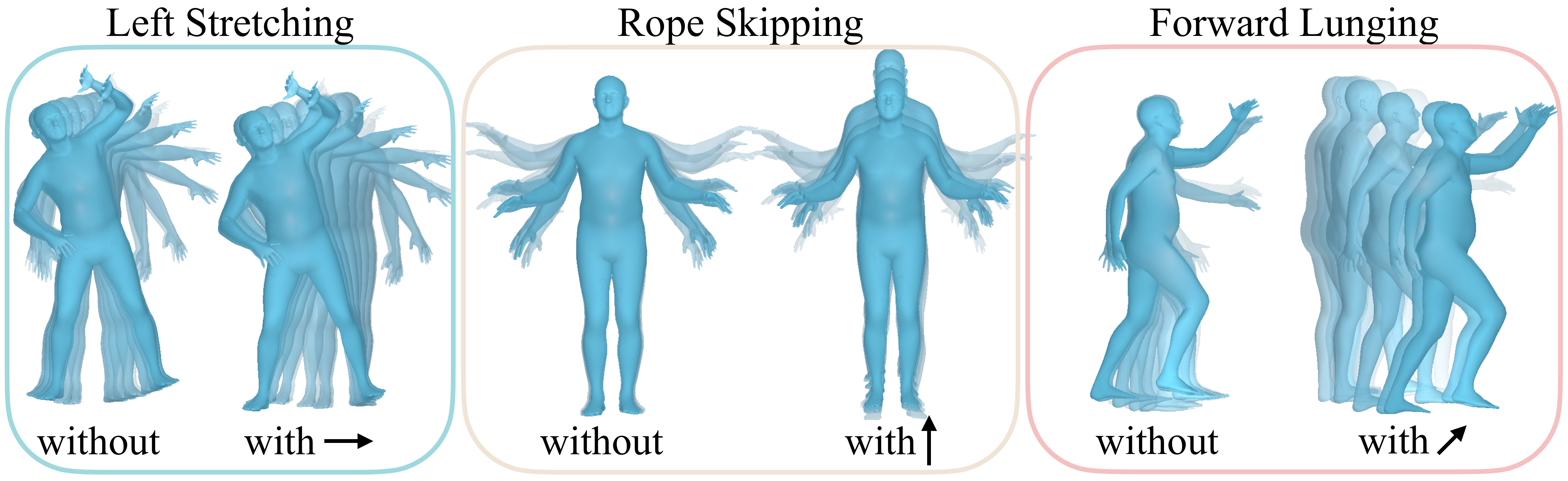}
    \vspace{-0.2cm}
    \caption{\textbf{Generating the 3D root translation:}
    Despite our model learning from noisy 3D trajectories,
    we show that our generations are smooth and they capture
    the semantics of the action. Examples are provided
    from the UESTC dataset for
    translations in $x$ (`Left Stretching'), $y$ (Rope Skipping),
    and $z$ (`Forward Lugging') with and without the loss on the
    root displacement $\mathcal{L}_{D}$.
    }
    \vspace{-0.3cm}
    \label{tab:translation}
\end{figure}

\noindent\textbf{Root translation.}
Since we estimate the 3D human body motion
from a monocular camera, obtaining the 3D trajectory
of the root joint is not trivial for real training sequences,
and is subject to depth
ambiguity. We assume a fixed focal length and approximate
the distance from the camera based on the ratio between
the 3D body height and the 2D projected height. Similar to
\cite{varol_surreact}, we observe reliable translation
in $xy$ image plane, but considerable noise in $z$ depth.
Nevertheless, we still train with this type of data and
visualize generated examples in Figure~\ref{tab:translation}
with and without the loss on translation $\mathcal{L}_{D}$.
Certain actions are defined by their trajectory (e.g., `Left Stretching')
and we are able to generate the semantically relevant
translations despite noisy data. Compared to the real
sequences, we observe much less noise in our generated sequences
(see the supplemental video at~\cite{projectpage}).

\noindent\textbf{Architecture design.}
Next, we ablate several architectural choices.
The first question is whether an attention-based
design (i.e., Transformer)
has advantages over the more widely used
alternatives such as a simple fully-connected
autoencoder or a GRU-based recurrent neural network.
In Table~\ref{tab:arch}, we see that 
our Transformer model outperforms both fully-connected
and GRU encoder-decoder architectures on two datasets
by a large margin.
\newedits{
In contrast to the fully-connected architecture,
we are also able to handle variable-length sequences.
We further note that our sequence-level decoding strategy is key to obtain
an improvement with Transformers, as opposed to an autoregressive Transformer decoder as in \cite{vaswani2017attention} (Table~\ref{tab:arch}, a). %
At training time, the autoregressive model uses teacher forcing,
i.e., using the ground-truth pose for the previous frame.
This creates a gap with test time, where we observed poor autoencoding reconstructions
such as decoding a left-hand waving encoding into a right-hand waving.}

We also provide a controlled
experiment by changing certain blocks of our Transformer VAE.
Specifically, we remove the $\mu_{a}^{token}$ and $\Sigma_{a}^{token}$
distribution parameter tokens and instead obtain $\mu$ and $\Sigma$ 
by averaging the outputs of the encoder, followed by two linear layers (Table~\ref{tab:arch}, b).
This results in considerable drop in performance.
Moreover, we investigate the additive $b_{a}^{token}$
token and
replace it with a one-hot encoding of the action label concatenated to the latent vector, followed by a linear projection (Table~\ref{tab:arch}, c).
\neweredits{Although this improves a bit} the results on the NTU-13 dataset,
we observe \neweredits{a large decrease in performance} on the UESTC dataset which has a larger number of action \neweredits{classes.} %

\begin{figure}
    \centering
    \includegraphics[width=0.49\linewidth]{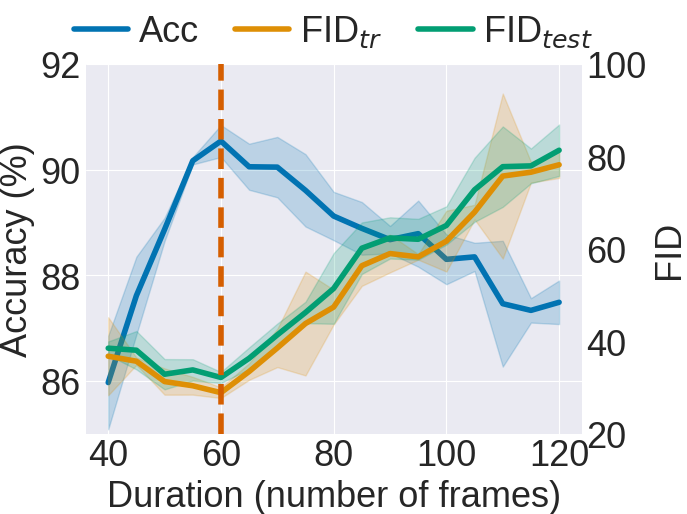}
    \includegraphics[width=0.49\linewidth]{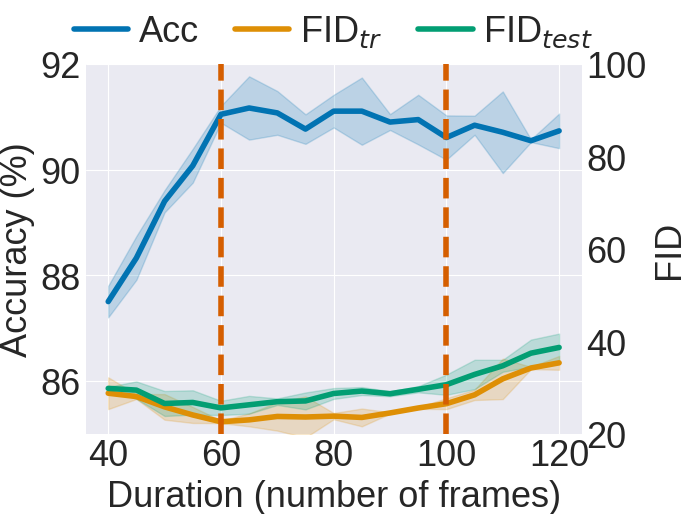}
    \vspace{-0.2cm}
    \caption{\textbf{Generating variable-length sequences:} We evaluate the capability
        of the models trained on UESTC with \textbf{(left) }fixed-size 60 frames and \textbf{(right)} variable-size between [60, 100] frames on generating various durations.
        We report accuracy and FID metrics.
        For the fixed model, we observe that the best performance
        is when tested at the seen duration of 60, but over 85\% accuracy
        is retained even at ranges between [40, 120] frames. The performance is overall improved
        when the model has previously seen duration variations
        in training; there is a smaller drop in performance
        beyond the seen range (denoted with dashed lines).
    }
    \label{fig:duration}
    \vspace{-0.3cm}
\end{figure}

\neweredits{Based on an architectural ablation of the number of Transformer
layers (see
\if\sepappendix1{Section~A.3}
\else{Section~\ref{app:subsec:numlayers}}
\fi
of the appendix),  we set this parameter to 8.}

\begin{table*}
    \centering
    \setlength{\tabcolsep}{6pt}
    \resizebox{0.99\linewidth}{!}{
    \begin{tabular}{lrrcc|rrcc}
        \toprule
        & \multicolumn{4}{c|}{NTU-13} & \multicolumn{4}{c}{HumanAct12} \\
        Method & \multicolumn{1}{c}{FID$_{tr}$$\downarrow$} & \multicolumn{1}{c}{Acc.$\uparrow$} & Div.$\rightarrow$ & Multimod.$\rightarrow$ & \multicolumn{1}{c}{FID$_{tr}$$\downarrow$} & \multicolumn{1}{c}{Acc.$\uparrow$} & Div.$\rightarrow$ & Multimod.$\rightarrow$ \\
        \midrule
        Real \cite{chuan2020action2motion} & $0.03^{\pm 0.00}$  & $99.9^{\pm 0.1}$  & $7.11^{\pm 0.05}$ & $2.19^{\pm 0.03}$
        & $0.09^{\pm 0.01}$ & $99.7^{\pm 0.1}$ & $6.85^{\pm 0.05}$ & $2.45^{\pm 0.04}$ \\
        Real*  & $0.02^{\pm0.00}$ & $99.8^{\pm0.0}$ & $7.07^{\pm0.02}$ & $2.25^{\pm0.01}$ & $0.02^{\pm0.00}$ & $99.4^{\pm0.0}$ & $6.86^{\pm0.03}$ & $2.60^{\pm0.01}$\\
        \midrule
        CondGRU~(\cite{chuan2020action2motion}$\dagger$) & $28.31^{\pm 0.14}$ & $7.8^{\pm 0.1}$ & $3.66^{\pm 0.02}$ & $3.58^{\pm 0.03}$ &
        $40.61^{\pm 0.14}$ & $8.0^{\pm 0.2}$ & $2.38^{\pm 0.02}$ & 
        $2.34^{\pm 0.04}$ \\
        Two-stage GAN~\cite{Cai_2018} (\cite{chuan2020action2motion}$\dagger$) & $13.86^{\pm 0.09}$ & $20.2^{\pm 0.3}$ & $5.33^{\pm 0.04}$ & $3.49^{\pm 0.03}$ & 
        $10.48^{\pm 0.09}$ & $42.1^{\pm 0.6}$ & $5.96^{\pm 0.05}$ & $2.81^{\pm 0.04}$ \\
        Act-MoCoGAN~\cite{Tulyakov:2018:MoCoGAN} (\cite{chuan2020action2motion}$\dagger$) & $2.72^{\pm 0.02}$ & ${\bf 99.7}^{\pm 0.1}$ & $6.92^{\pm 0.06}$ & $0.91^{\pm 0.01}$ & 
        $5.61^{\pm 0.11}$ & $79.3^{\pm 0.4}$ & $6.75^{\pm 0.07}$ & 
        $1.06^{\pm 0.02}$ \\
        Action2Motion~\cite{chuan2020action2motion} & $0.33^{\pm 0.01}$  & $94.9^{\pm 0.1}$  & $7.07^{\pm 0.04}$ & $2.05^{\pm 0.03}$ & 
        $2.46^{\pm 0.08}$ & $92.3^{\pm 0.2}$ & $7.03^{\pm 0.04}$ & 
        $2.87^{\pm 0.04}$ \\
        \midrule
        \rowcolor{aliceblue}
        ACTOR (ours) & ${\bf 0.11}^{\pm0.00}$ & $97.1^{\pm0.2}$ & $7.08^{\pm0.04}$ & $2.08^{\pm0.01}$ &${\bf0.12}^{\pm0.00}$ & ${\bf 95.5}^{\pm0.8}$ & $6.84^{\pm0.03}$ & $2.53^{\pm0.02}$\\
        \bottomrule
    \end{tabular}
    }
    \vspace{-0.3cm}
    \caption{\textbf{State-of-the-art comparison:}
        We compare to the recent work of \cite{chuan2020action2motion}
        on the NTU-13 and HumanAct12 datasets. Note that due to differences in implementation
        (e.g., random sampling, using zero shape parameter),
        our metrics for the ground truth real data (Real*) are slightly different
        than the ones reported in their paper. The performance improvement
        with our Transformer-based model shows a clear gap from Action2Motion.
        $\dagger$ Baselines implemented by \cite{chuan2020action2motion}.
    }
    \vspace{-0.3cm}
    \label{tab:sota}
\end{table*}

\noindent\textbf{Training with sequences of variable durations.}
A key advantage of sequence-modeling with architectures
such as Transformers is to be able to handle variable-length
motions. At generation time, we control how long
the model should synthesize, by specifying a sequence
of positional encodings to the decoder.
We can trivially generate more diversity by synthesizing
sequences of different durations. However,
so far we have trained our models with fixed-size
inputs, i.e., 60 frames. Here, we first analyze whether
a fixed-size trained model can directly generate variable sizes.
This is presented in Figure~\ref{fig:duration} (left).
We plot the performance over several sets of generations
of different lengths between 40 and 120 frames (with a step size of 5).
Since our recognition model used for evaluation metrics
is trained on fixed-size 60-frame inputs, we naturally observe
performance decrease outside of this length. However,
the accuracy still remains high which indicates
that our model is already capable of generating diverse durations.

Next, we \textit{train} our generative model
with variable-length inputs by randomly sampling a sequence
between 60 and 100 frames. However, simply training
this way from random weight initialization converges to a poor solution,
leading all generated motions to be frozen in time.
We address this by pretraining at 60-frame fixed size and finetuning at variable sizes.
We see in Figure~\ref{fig:duration} (right)
that the performance is greatly improved with this model.

Furthermore, we investigate how the generations longer or shorter
than their average durations behave. We observe qualitatively that
shorter generations produce 
\newedits{partial actions}
e.g.,
picking up without reaching the floor, and
longer generations slow down somewhat non-uniformly in time.
We refer to the supplemental video  %
\cite{projectpage} for qualitative results.

\subsection{Comparison to the state of the art}
\label{subsec:sota}

Action2Motion~\cite{chuan2020action2motion} is the only prior work that
generates action-conditioned motions.
We compare to them in Table~\ref{tab:sota}
on their NTU-13 and HumanAct12 datasets.
On both datasets, we obtain significant improvements
over this prior work that uses
autoregressive GRU blocks, as well as other baselines
implemented by \cite{chuan2020action2motion} by adapting
other works \cite{Cai_2018,Tulyakov:2018:MoCoGAN}.
\newedits{The improvements over \cite{chuan2020action2motion} can be explained mainly
by removing autoregression and adding the proposed learnable tokens (Table~\ref{tab:arch}).
Note that our GRU implementation obtains similar performance as \cite{chuan2020action2motion}, while using
the same hyperparameters as the Transformer.}
In addition to the quantitative performance improvement,
measured with recognition models based on joint coordinates,
our model \neweredits{can directly output human meshes},
which can further be diversified with varying the shape parameters. 
\cite{chuan2020action2motion} instead applies
an optimization step to fit SMPL models on their
generated joint coordinates, which is typically
substantially
slower than a neural network forward pass.

\begin{table}
    \centering
    \setlength{\tabcolsep}{6pt}
    \resizebox{0.8\linewidth}{!}{
    \begin{tabular}{l|cc}
        \toprule
        & \multicolumn{2}{c}{Test accuracy (\%)} \\
        & Real$_{orig}$ & Real$_{denoised}$ \\
        \midrule
        Real$_{orig}$ & 91.8 & 93.2 \\
        Real$_{denoised}$ & 83.8 & 97.0 \\
        \midrule
        Real$_{interpolated}$ & 77.6 & 93.9\\
        Generated & 80.7 & 97.0 \\
        Real$_{orig}$ + Generated & \textbf{91.9} & \textbf{98.3} \\
        \bottomrule
    \end{tabular}
    }
    \vspace{-0.3cm}
    \caption{\textbf{Action recognition:} We employ a standard architecture
            (ST-GCN~\cite{stgcn}) and perform action recognition experiments using
            several sets of training data on the UESTC cross-subject protocol~\cite{uestc2018}.
            Training only with generated samples obtains 80\% accuracy on the real
            test set
            which is another indication our action-conditioning performs well.
            Nevertheless, we observe a domain gap between generated and real samples,
            mainly due to the noise present in the real data. We show that
            simply by encoding-decoding the test sequences, we observe
            a denoising effect, which in turn shows better performance. However,
            one should note that the last-column experiments are not meant to improve performance
            in the benchmark since it uses the action label information.
    }
    \vspace{-0.3cm}
    \label{tab:recognition}
\end{table}

\begin{figure*}[t]
    \centering
    \includegraphics[width=0.9\linewidth]{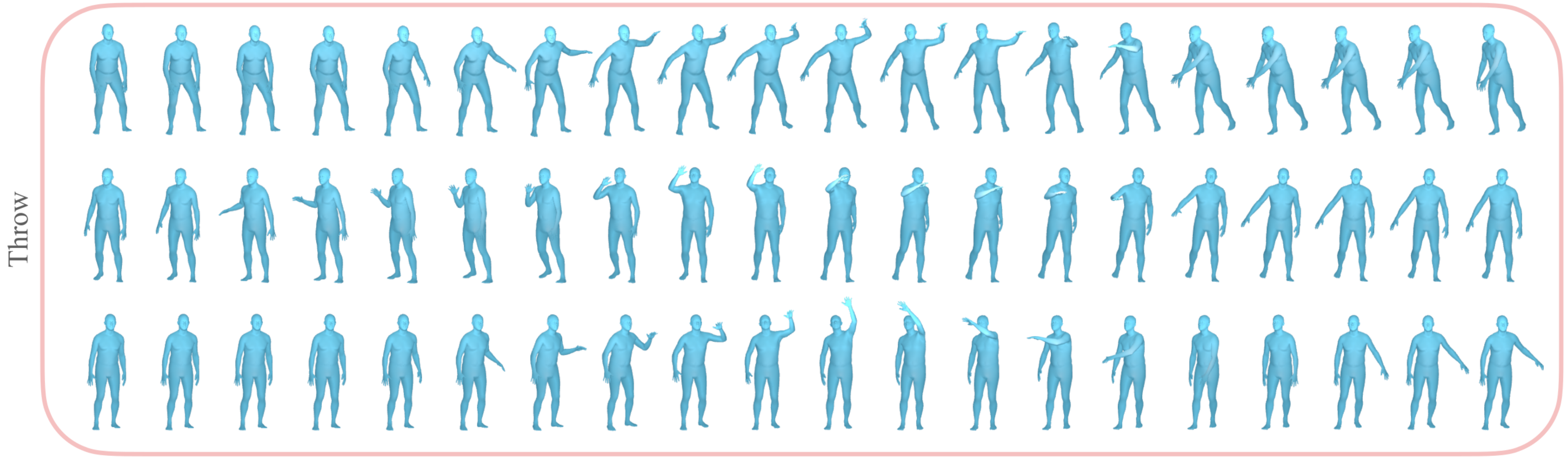}
    \vspace{-0.2cm}
    \caption{\textbf{Qualitative results:} We illustrate
        the diversity of our generations %
        on the `Throw' action from NTU-13 by showing 3 sequences.
        The horizontal axis represent the time axis
        and 20 equally spaced frames are visualized out of 60-frame generations.
        We demonstrate that our model is capable of generating
        different ways of performing a given action. More results
        can be found in
        \if\sepappendix1{Section~B}
        \else{Section~\ref{app:sec:qualitative}}
        \fi
        of the appendix
        and the supplemental video at~\cite{projectpage}.
    }
    \label{fig:qualitative}
    \vspace{-0.4cm}
\end{figure*}

\begin{figure}
    \centering
    \includegraphics[width=0.59\linewidth]{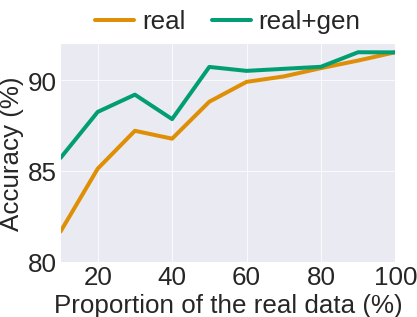}
    \vspace{-0.3cm}
    \caption{\textbf{Data augmentation:}
        We show the benefit of augmenting the
        real data with our generative model (real+gen), especially
        at low-data regime. We have limited gains when the real
        data is sufficiently large.}
    \vspace{-0.5cm}
    \label{fig:recognition}
\end{figure}

\subsection{Use cases in action recognition}
\label{subsec:usecases}
In this section, we test the limits of our approach
by illustrating the benefits
of our generative model and our learned latent representation
for the skeleton-based action recognition task.
We adopt a standard architecture, ST-GCN~\cite{stgcn},
that employs spatio-temporal graph convolutions
to classify actions. We show that we can use
our latent encoding for denoising
motion estimates and our generated sequences as
data augmentation to action recognition models.

\noindent\textbf{Use case I: Human motion denoising.}
In the case when our motion data source relies on
monocular motion estimation such as \cite{VIBECVPR2020},
the training motions remain noisy. We observe that
by simply encoding-decoding the real motions
through our learned embedding space, we obtain
much cleaner motions.
Since it is difficult to show motion quality results
on static figures, we refer to our supplemental video at \cite{projectpage}
to see this effect. We measure the denoising capability
of our model through an action recognition experiment in Table~\ref{tab:recognition}.
We change both the training and test set motions
with the encoded-decoded versions. We show
improved performance when trained and tested on
Real$_{denoised}$ motions (97.0\%) compared to Real$_{orig}$ (91.8).
Note that this result on its own is not sufficient
for this claim, but is only an indication
since our decoder might produce less diversity than real motions.
Moreover, the action label is given at denoising time.
We believe that such denoising can be beneficial
in certain scenarios where the action is known, e.g.,
occlusion or missing markers during MoCap collection.

\noindent\textbf{Use case II: Augmentation for action recognition.}
Next, we augment the real
training data (Real$_{orig}$), by adding generated motions to the training.
We first measure the action recognition performance without using
real sequences.
We consider interpolating existing Real$_{orig}$ motions
that fall within the same action category in our embedding space
to create intra-class variations 
(Real$_{interpolated}$). 
We then synthesize
motions by sampling noise vectors conditioned on each action category (Generated).
Table~\ref{tab:recognition} summarizes the results.
Training only on synthetic %
data
performs 80.7\% on the real test set, which is promising.
However, there is a domain gap between the noisy real motions
and our smooth generations.
Consequently, adding generated motions to real training
only marginally improves the performance.
In Figure~\ref{fig:recognition},
we investigate whether the augmented training helps
for low-data regimes by training at several fractions of the data.
In each minibatch we equally sample real and generated motions.
However, in theory we have access to infinitely many generations.
We see that the improvement is more visible at low-data regime.

\subsection{Qualitative results}
\label{subsec:qualitative}

In Figure~\ref{fig:qualitative},
we visualize several examples from our generations.
We observe a great diversity
in the way a given action is performed.
For example, the `Throw' action is performed
with left or right hand.
\newedits{We notice that the model keeps the essence
of the action semantics while changing nuances
(angles, speed, phase) or action-irrelevant body parts.}
We refer to 
the supplemental video at~\cite{projectpage} and
\if\sepappendix1{Section~B}
\else{Section~\ref{app:sec:qualitative}}
\fi
of the appendix
for further qualitative analyses.

One limitation of our model is that the maximum
duration it can generate depends on computational resources
since we output all the sequence at once. Moreover,
the actions are from a predefined
set. Future work will explore open-vocabulary actions,
which might become possible with further progress in 3D motion
estimation from unconstrained videos.

\section{Conclusions}
\label{sec:conclusions}

We presented a new Transformer-based
VAE model to synthesize action-conditioned
human motions. We provided a detailed
analysis to assess different components
of our proposed approach. 
We obtained state-of-the-art
performance on action-conditioned motion generation,
significantly improving over prior work.
Furthermore, we explored various use cases
in motion denoising and action recognition.
One especially attractive property
of our method is that it operates on a sequence-level
latent space.
Future work can therefore exploit our model
to impose priors on motion estimation
or action recognition problems.

\bigskip

{\footnotesize
\noindent\textbf{Acknowledgements.}
This work was granted access to
the HPC resources of IDRIS under the allocation 2021-101535 made by GENCI.
The authors would like to thank Mathieu Aubry and David Picard
for helpful feedback, Chuan Guo and Shihao Zou for their help with Action2Motion details.
\par}

{\footnotesize
\noindent\textbf{Disclosure:} MJB has received research funds
from Adobe, Intel, Nvidia, Facebook, and Amazon. While MJB
is a part-time employee of Amazon, his research was performed
solely at, and funded solely by, Max Planck. MJB has financial interests in Amazon, Datagen Technologies, and Meshcapade
GmbH.
\par
}

{\small
\bibliographystyle{ieee_fullname}
\bibliography{references}
}

\bigskip
{\noindent \large \bf {APPENDIX}}\\
\renewcommand{\thefigure}{A.\arabic{figure}} %
\setcounter{figure}{0} 
\renewcommand{\thetable}{A.\arabic{table}}
\setcounter{table}{0} 

\appendix

\newcommand{\vspaceifappendix}[1]{\if\sepappendix1{\vspace{#1}} \fi}

\if\sepappendix1{\vspace{-0.3cm}
This document provides
additional experiments (Section~\ref{app:sec:experiments}),
additional qualitative results 
(Section~\ref{app:sec:qualitative}),
and implementation details (Section~\ref{app:sec:implementation}).}
\else{We provides
additional experiments (Section~\ref{app:sec:experiments}),
additional qualitative results 
(Section~\ref{app:sec:qualitative}),
and implementation details (Section~\ref{app:sec:implementation}).}
\fi

\vspaceifappendix{-0.2cm}  %
\section{Additional experiments}
\label{app:sec:experiments}
\vspaceifappendix{-0.10cm}

We ablate the model and vary key parameters to evaluate
the influence of  design choices
on the quality of the results. In particular, we present
results on the effect of
$\lambda_{KL}$ (Section~\ref{app:subsec:kl}),
batch size (Section~\ref{app:subsec:batchsize}),
number of layers (Section~\ref{app:subsec:numlayers}),
and the
rotation representation for SMPL pose parameters (Section~\ref{app:subsec:poserep}).

\vspaceifappendix{-0.2cm}
\subsection{Weight of the KL loss}
\label{app:subsec:kl}
\vspaceifappendix{-0.15cm}

As explained in
\if\sepappendix1{Section~3.2}
\else{Section~\ref{subsec:training}}
\fi
of the main paper,
we empirically show the importance of the weighting
between the reconstruction loss and the KL loss,
controlled by $\lambda_{KL}$.
Table~\ref{app:tab:kl} presents results for
several values of $\lambda_{KL}$ and we find that
there is a trade-off between diversity and realism
that is best balanced at $\lambda_{KL}=1\mathrm{e}{-5}$.
We use this value in all our experiments.

\vspaceifappendix{-0.2cm}
\subsection{Influence of the batch size}
\label{app:subsec:batchsize}
\vspaceifappendix{-0.15cm}

As pointed out in
\if\sepappendix1{Section~3.2}
\else{Section~\ref{subsec:training}}
\fi
of the main paper,
we find that the batch size significantly
influences the performance.
In Table~\ref{app:tab:batchsize},
we report results with batch sizes of 10, 20, 30, 40
for a fixed learning rate. The best performance
is obtained at 20, which is used in all our experiments.

\vspaceifappendix{-0.2cm}
\subsection{Number of layers}
\label{app:subsec:numlayers}
\vspaceifappendix{-0.15cm}

We experiment with the number of Transformer layers
in both of our encoder and decoder architectures.
Table~\ref{app:tab:numlayers} summarizes the results.
While 2 and 4 layers are sub-optimal, the performance
difference between 6 and 8 layers is minimal.
We use 8 layers in all our experiments.

\vspaceifappendix{-0.2cm}
\subsection{SMPL pose parameter representation}
\label{app:subsec:poserep}
\vspaceifappendix{-0.15cm}

In Table~\ref{app:tab:poserep},
we explore different rotation representations
for SMPL pose parameters.
Note that we also preserve
the loss on the vertices $\mathcal{L}_{V}$ in all rows.
We find that an axis-angle representation
is difficult to train due to discontinuities, while others, such as
quaternions, rotation matrices and 6D continuous representations
\cite{Zhou2019OnTC}
are similar in performance on NTU-13. On UESTC,
we obtain the best performance
with the 6D representation and use this in all our experiments.

\vspaceifappendix{-0.1cm}
\section{Additional qualitative results}
\vspaceifappendix{-0.1cm}
\label{app:sec:qualitative}

\begin{figure*}[h!]
    \centering
    \includegraphics[width=.99\linewidth]{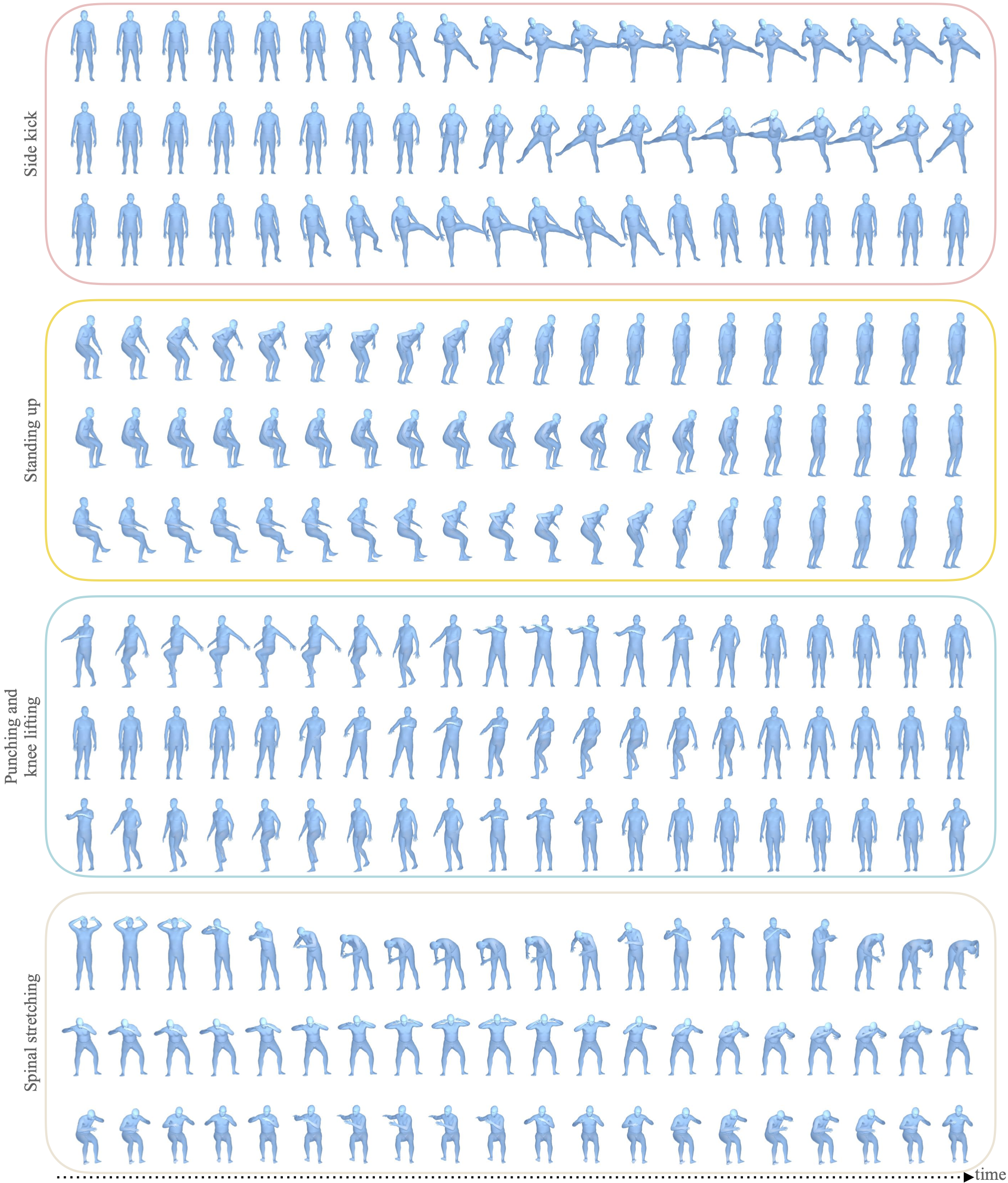}
    \caption{\textbf{Additional qualitative results:}
        We provide more action categories from NTU-13 (top two actions: `Side kick'
        and `Standing up')
        and UESTC (bottom two actions: `Punching and knee lifting' and `Spinal stretching').
        As in
    \if\sepappendix1{Figure~6}
    \else{Section~\ref{fig:qualitative}}
    \fi
    of the main paper, we show 3 generations per action. Our model
    generates different ways to perform the same action.
    }
    \label{app:fig:qualitative}
\end{figure*}

Figure~\ref{app:fig:qualitative}
demonstrates the diversity of our generated
motions for additional actions on NTU-13 and UESTC.

\noindent\textbf{Video.}
We provide a 
supplemental video at~\cite{projectpage}
to illustrate qualitatively the diversity
in our generations and compare with 
Action2Motion~\cite{chuan2020action2motion}. Moreover, we 
visualize the effect
of using a combined reconstruction loss defined
both on rotations and vertex coordinates, as opposed
to a single loss.
We further present results of changing the duration
of the generations. We also inspect
the latent space by interpolating the noise vector.
Finally, we present
the denoising capability of our model by
encoding-decoding through our latent space.
This takes jerky motions and produces smooth but natural looking motion.

\noindent\newedits{\textbf{Jitter removal for Action2Motion~\cite{chuan2020action2motion}.}
Besides the quantitative improvement of ACTOR over Action2Motion,
we observe qualitatively that Action2Motion generations have
significant temporal jitter. To investigate whether our improvement
stems from this difference, we removed jitter (using 1\euro{} filter)
from Action2Motion generations (that we obtained with their code).
The result becomes worse (FID: 0.41 $\rightarrow$ 0.63, Acc: 94.3\% $\rightarrow$ 93.0\%)\footnote{\newedits{These two values for Action2Motion does not match Table 3 of the paper because we use our own evaluation script and normalized the ground truth shape by taking the average shape of SMPL.}},
perhaps because the real data also has considerable jitter. This suggests that
our significant quantitative improvement 
can be attributed to other factors such as more distinguishable actions.}

\vspaceifappendix{-0.1cm}
\section{Implementation details}
\label{app:sec:implementation}
\vspaceifappendix{-0.1cm}

\noindent\textbf{Architectural details.} For all our experiments, we set the embedding dimensionality to 256. In the Transformer, we set the number of layers to 8, the number of heads in multi-head attention to 4, the dropout rate to 0.1 and the dimension of the intermediate feedforward network to 1024.
As in GPT-3~\cite{brown2020language} and BERT~\cite{devlin2018bert}, we use Gaussian Linear Error Units (GELU)~\cite{hendrycks2016gaussian} in our Transformer architecture.

\vspaceifappendix{0.1cm}

\noindent\textbf{Library credits.}
Our models are implemented with PyTorch~\cite{NEURIPS2019pytorch}, 
and we use PyTorch3D~\cite{ravi2020pytorch3d} to perform 
differentiable conversion between rotation representations.
We integrate the differentiable SMPL layer using the PyTorch 
implementation of SMPL-X~\cite{SMPL-X:2019}.

\vspaceifappendix{0.1cm}

\noindent\textbf{Metrics.}
For the evaluation metrics, we use the implementations
provided by Action2Motion~\cite{chuan2020action2motion}.

\vspaceifappendix{0.1cm}

\noindent\textbf{Runtime.} Training takes 24 hours for 2K epochs on NTU,
19h hours for 5K epochs on HumanAct12,
and 33 hours for 1K epochs on UESTC on a single Tesla V100 GPU, using 4GB GPU memory
with batch size 20.

\vspaceifappendix{0.1cm}

\noindent\textbf{Training  with  sequences  of  variable  durations.}
As explained in
\if\sepappendix1{Section~4.2}
\else{Section~\ref{subsec:ablations}}
\fi
of the main paper,
we finetune our model with variable-durations after pretraining on fixed-durations.
For this, we restore the model weights from the fixed-duration
pretraining and finetune for 100 additional epochs, with the same training hyperparameters.

\begin{table*}
    \centering
    \setlength{\tabcolsep}{6pt}
    \resizebox{0.99\linewidth}{!}{
    \begin{tabular}{lrrccr|rccc}
        \toprule
        & \multicolumn{5}{c}{UESTC} & \multicolumn{4}{|c}{NTU-13} \\
     & \multicolumn{1}{c}{FID$_{tr}$$\downarrow$} & \multicolumn{1}{c}{FID$_{test}$$\downarrow$} & Acc.$\uparrow$ & Div.$\rightarrow$ & \multicolumn{1}{c}{Multimod.$\rightarrow$} & \multicolumn{1}{c}{FID$_{tr}$$\downarrow$} & Acc.$\uparrow$ & Div.$\rightarrow$ & Multimod.$\rightarrow$ \\
        \midrule
        Real & $2.93^{\pm0.26}$ & $2.79^{\pm0.29}$ & $98.8^{\pm0.1}$ & $33.34^{\pm0.32}$ & $14.16^{\pm0.06}$ & $0.02^{\pm0.00}$ & $99.8^{\pm0.0}$ & $7.07^{\pm0.02}$ & $2.27^{\pm0.01}$\\
        \midrule
        $\lambda_{KL}=1\mathrm{e}{-3}$ & $460.72^{\pm90.36}$ & $490.12^{\pm36.10}$ & $34.4^{\pm1.4}$ & $20.69^{\pm0.60}$ & $1.25^{\pm0.00}$ & $13.79^{\pm0.03}$ & $46.6^{\pm0.7}$ & $5.79^{\pm0.04}$ & $1.53^{\pm0.01}$\\
        $\lambda_{KL}=1\mathrm{e}{-4}$ & $367.95^{\pm94.07}$ & $390.68^{\pm41.02}$ & $38.1^{\pm0.9}$ & $20.91^{\pm0.38}$ & $9.19^{\pm0.08}$ & $9.90^{\pm0.02}$ & $50.3^{\pm1.0}$ & $6.15^{\pm0.04}$ & $2.86^{\pm0.02}$\\
        \rowcolor{aliceblue}
        $\lambda_{KL}=1\mathrm{e}{-5}$ & $20.02^{\pm1.79}$ & $23.64^{\pm3.59}$ & $90.5^{\pm0.4}$ & $32.77^{\pm0.48}$ & $14.64^{\pm0.07}$ & $0.17^{\pm0.00}$ & $96.4^{\pm0.2}$ & $7.08^{\pm0.03}$ & $2.12^{\pm0.01}$\\
        $\lambda_{KL}=1\mathrm{e}{-6}$ & $34.13^{\pm5.52}$ & $39.74^{\pm3.57}$ & $77.4^{\pm0.8}$ & $29.60^{\pm0.35}$ & $18.08^{\pm0.08}$ & $13.83^{\pm0.03}$ & $46.6^{\pm0.7}$ & $5.78^{\pm0.04}$ & $1.54^{\pm0.01}$\\
        $\lambda_{KL}=1\mathrm{e}{-7}$ & $80.05^{\pm7.71}$ & $83.68^{\pm12.55}$ & $47.1^{\pm2.1}$ & $25.06^{\pm0.15}$ & $19.96^{\pm0.08}$ & $7.04^{\pm0.03}$ & $43.0^{\pm2.1}$ & $6.17^{\pm0.03}$ & $4.18^{\pm0.01}$\\
        \bottomrule
    \end{tabular}
    }
    \caption{\textbf{Weighting the KL loss term:}
        To obtain a good trade-off between diversity and realism,
        it is important to find the balance between the reconstruction
        loss term and the KL loss term in training. We set the weight
        $\lambda_{KL}$ to $1\mathrm{e}{-5}$ in our training.
    }
    \label{app:tab:kl}
\end{table*}
\begin{table*}
    \centering
    \setlength{\tabcolsep}{6pt}
    \resizebox{0.99\linewidth}{!}{
    \begin{tabular}{lccccc|cccc}
        \toprule
        & \multicolumn{5}{c}{UESTC} & \multicolumn{4}{|c}{NTU-13} \\
        & FID$_{tr}$$\downarrow$ & FID$_{test}$$\downarrow$ & Acc.$\uparrow$ & Div.$\rightarrow$ & Multimod.$\rightarrow$ & FID$_{tr}$$\downarrow$ & Acc.$\uparrow$ & Div.$\rightarrow$ & Multimod.$\rightarrow$ \\
        \midrule
        Real & $2.93^{\pm0.26}$ & $2.79^{\pm0.29}$ & $98.8^{\pm0.1}$ & $33.34^{\pm0.32}$ & $14.16^{\pm0.06}$ & $0.02^{\pm0.00}$ & $99.8^{\pm0.0}$ & $7.07^{\pm0.02}$ & $2.27^{\pm0.01}$\\
        \midrule
        Batch size = 10 & $283.28^{\pm94.40}$ & $309.15^{\pm33.90}$ & $39.7^{\pm1.5}$ & $23.24^{\pm0.43}$ & $15.73^{\pm0.11}$ & $13.95^{\pm0.03}$ & $46.2^{\pm0.6}$ & $5.77^{\pm0.05}$ & $1.56^{\pm0.01}$\\
        \rowcolor{aliceblue}
        Batch size = 20 & $20.02^{\pm1.79}$ & $23.64^{\pm3.59}$ & $90.5^{\pm0.4}$ & $32.77^{\pm0.48}$ & $14.64^{\pm0.07}$ & $0.17^{\pm0.00}$ & $96.4^{\pm0.2}$ & $7.08^{\pm0.03}$ & $2.12^{\pm0.01}$\\
        Batch size = 30 & $23.37^{\pm2.95}$ & $26.06^{\pm1.28}$ & $89.7^{\pm0.5}$ & $32.07^{\pm0.58}$ & $14.59^{\pm0.05}$ & $0.18^{\pm0.00}$ & $96.2^{\pm0.2}$ & $7.07^{\pm0.04}$ & $2.13^{\pm0.01}$\\
        Batch size = 40 & $25.36^{\pm1.82}$ & $28.22^{\pm2.16}$ & $89.2^{\pm0.7}$ & $32.22^{\pm0.44}$ & $14.52^{\pm0.10}$ & $0.26^{\pm0.00}$ & $95.4^{\pm0.1}$ & $7.06^{\pm0.05}$ & $2.10^{\pm0.01}$\\
        \bottomrule
    \end{tabular}
    }
    \caption{\textbf{Batch size:}
        We observe sensitivity of the Transformer VAE training
        to different batch sizes and report performances at several
        batch size values. We set this hyperparameter to 20 in our training.
    }
    \label{app:tab:batchsize}
\end{table*}
\begin{table*}
    \centering
    \setlength{\tabcolsep}{6pt}
    \resizebox{0.99\linewidth}{!}{
    \begin{tabular}{lccccc|cccc}
        \toprule
        & \multicolumn{5}{c}{UESTC} & \multicolumn{4}{|c}{NTU-13} \\
        & FID$_{tr}$$\downarrow$ & FID$_{test}$$\downarrow$ & Acc.$\uparrow$ & Div.$\rightarrow$ & Multimod.$\rightarrow$ & FID$_{tr}$$\downarrow$ & Acc.$\uparrow$ & Div.$\rightarrow$ & Multimod.$\rightarrow$ \\
        \midrule
        Real & $2.93^{\pm0.26}$ & $2.79^{\pm0.29}$ & $98.8^{\pm0.1}$ & $33.34^{\pm0.32}$ & $14.16^{\pm0.06}$ & $0.02^{\pm0.00}$ & $99.8^{\pm0.0}$ & $7.07^{\pm0.02}$ & $2.27^{\pm0.01}$\\
        \midrule
        2-layers & $34.66^{\pm2.58}$ & $37.17^{\pm3.53}$ & $84.9^{\pm0.6}$ & $30.87^{\pm0.36}$ & $15.83^{\pm0.08}$ & $0.24^{\pm0.00}$ & $94.6^{\pm0.2}$ & $7.07^{\pm0.03}$ & $2.22^{\pm0.01}$\\
        4-layers & $23.93^{\pm1.50}$ & $26.75^{\pm1.99}$ & $88.9^{\pm0.5}$ & $32.24^{\pm0.76}$ & $15.06^{\pm0.06}$ & $0.19^{\pm0.00}$ & $96.1^{\pm0.2}$ & $7.09^{\pm0.04}$ & $2.10^{\pm0.01}$\\
        6-layers & $21.68^{\pm1.78}$ & $24.92^{\pm2.09}$ & $89.0^{\pm0.6}$ & $32.61^{\pm0.41}$ & $15.31^{\pm0.05}$ & $0.16^{\pm0.00}$ & $96.6^{\pm0.1}$ & $7.09^{\pm0.04}$ & $2.11^{\pm0.01}$\\
        \rowcolor{aliceblue}
        8-layers & $20.02^{\pm1.79}$ & $23.64^{\pm3.59}$ & $90.5^{\pm0.4}$ & $32.77^{\pm0.48}$ & $14.64^{\pm0.07}$ & $0.17^{\pm0.00}$ & $96.4^{\pm0.2}$ & $7.08^{\pm0.03}$ & $2.12^{\pm0.01}$\\
        \bottomrule
    \end{tabular}
    }
    \caption{\textbf{Number of layers:}
        We use 8 layers in both the encoder and the decoder
        of the Transformer VAE. While the performance degrades
        at 2 or 4 layers, we see marginal gains after 6 layers.
    }
    \label{app:tab:numlayers}
\end{table*}
\begin{table*}
    \centering
    \setlength{\tabcolsep}{6pt}
    \resizebox{0.99\linewidth}{!}{
     \begin{tabular}{lccccr|cccc}
        \toprule
        & \multicolumn{5}{c}{UESTC} & \multicolumn{4}{|c}{NTU-13} \\
         & FID$_{tr}$$\downarrow$ & FID$_{test}$$\downarrow$ & Acc.$\uparrow$ & Div.$\rightarrow$ & \multicolumn{1}{c}{Multimod.$\rightarrow$} & FID$_{tr}$$\downarrow$ & Acc.$\uparrow$ & Div.$\rightarrow$ & Multimod.$\rightarrow$ \\
        \midrule
        Real & $2.93^{\pm0.26}$ & $2.79^{\pm0.29}$ & $98.8^{\pm0.1}$ & $33.34^{\pm0.32}$ & $14.16^{\pm0.06}$ & $0.02^{\pm0.00}$ & $99.8^{\pm0.0}$ & $7.07^{\pm0.02}$ & $2.27^{\pm0.01}$\\
        \midrule
        Axis-angle & $513.39^{\pm98.35}$ & $531.88^{\pm43.41}$ & $16.4^{\pm0.4}$ & $19.75^{\pm0.44}$ & $1.81^{\pm0.00}$ & $14.98^{\pm0.03}$ & $41.7^{\pm0.7}$ & $5.29^{\pm0.02}$ & $1.96^{\pm0.01}$\\
        Quaternion & $281.9^{\pm87.5}$ & $305.02^{\pm21.97}$ & $41.2^{\pm1.0}$ & $23.48^{\pm0.39}$ & $14.57^{\pm0.06}$ & $0.20^{\pm0.00}$ & $95.6^{\pm0.3}$ & $7.08^{\pm0.04}$ & $2.23^{\pm0.01}$\\
        Rotation matrix & $277.14^{\pm76.59}$ & $300.29^{\pm29.53}$ & $41.6^{\pm1.9}$ & $22.25^{\pm0.30}$ & $14.56^{\pm0.10}$ & $0.17^{\pm0.00}$ & $95.9^{\pm0.2}$ & $7.08^{\pm0.04}$ & $2.19^{\pm0.01}$\\
        \rowcolor{aliceblue}
    6D continuous & $20.02^{\pm1.79}$ & $23.64^{\pm3.59}$ & $90.5^{\pm0.4}$ & $32.77^{\pm0.48}$ & $14.64^{\pm0.07}$ & $0.17^{\pm0.00}$ & $96.4^{\pm0.2}$ & $7.08^{\pm0.03}$ & $2.12^{\pm0.01}$\\
        \bottomrule
    \end{tabular}
    }
    \caption{\textbf{SMPL pose parameter representation:}
        We investigate different rotation representations for the SMPL
        pose parameters. While on NTU-13, all except axis-angle representations
        perform similarly, the best performing representation on UESTC is the
        6D continuous representation~\cite{Zhou2019OnTC}. Note that
        the action recognition model which is used for evaluation
        is based on 6D rotations on UESTC and joint coordinates on NTU-13.
        Therefore, we convert each generation to these representations before evaluation.
    }
    \label{app:tab:poserep}
\end{table*}

\end{document}